%% file: main.tex
\documentclass{article}

\usepackage{microtype}
\usepackage{graphicx}
\usepackage{subcaption}
\usepackage{booktabs} 

\usepackage{hyperref}

\usepackage[preprint]{icml2026}

\usepackage{amsmath}
\usepackage{amssymb}
\usepackage{mathtools}
\usepackage{amsthm}

\usepackage[capitalize,noabbrev]{cleveref}

\newcommand{\ignore}[1]{}
\newcommand{\authorcomment}[3]{\textcolor{#1}{[#3: \textsf{#2}]}}

\newcommand{\ag}[1]{\authorcomment{brown}{#1}{AG}}
\newcommand{\mg}[1]{\authorcomment{green}{#1}{MG}}
\newcommand{\ls}[1]{\authorcomment{orange}{#1}{LS}}

\renewcommand{\authorcomment}[3]{}

\newcommand{\tb}{$T$\xspace}
\newcommand{\cb}{$C$\xspace}
\newcommand{\tbm}{T}
\newcommand{\cbm}{C}

\newcommand{\llm}{$M_\theta$\xspace}

\theoremstyle{plain}

\theoremstyle{definition}

\theoremstyle{remark}

\usepackage[textsize=tiny]{todonotes}

\usepackage{enumitem} 
\usepackage{xspace}

\usepackage[skins, breakable]{tcolorbox}

\newcommand{\arch}{Thinking States\xspace}

\icmltitlerunning{Latent Reasoning with Supervised Thinking States}

\begin{document}

\twocolumn[
  \icmltitle{Latent Reasoning with Supervised Thinking States}



    \icmlsetsymbol{intern}{*}


    \begin{icmlauthorlist}
        \icmlauthor{Ido Amos}{google,huji,intern}
        \icmlauthor{Avi Caciularu}{google}
        \icmlauthor{Mor Geva}{google,tau}
        \icmlauthor{Amir Globerson}{google,tau}
        \icmlauthor{Jonathan Herzig}{google}
        \icmlauthor{Lior Shani}{google}
        \icmlauthor{Idan Szpektor}{google}
    \end{icmlauthorlist}

    \icmlaffiliation{google}{Google Research}
    \icmlaffiliation{huji}{The Hebrew University of Jerusalem}
    \icmlaffiliation{tau}{Tel Aviv University}

  \icmlcorrespondingauthor{Ido Amos}{idoamos@mail.tau.ac.il}

  \icmlkeywords{Machine Learning, ICML}

  \vskip 0.3in
]



\printAffiliationsAndNotice{\textsuperscript{*}Work done during an internship at Google Research.}

\begin{abstract}
Reasoning with a chain-of-thought (CoT) enables Large Language Models (LLMs) to solve complex tasks but incurs significant inference costs due to the generation of long rationales.
We propose Thinking States, a method that performs reasoning {\em while} the input is processing. Specifically, Thinking States generates sequences of thinking tokens every few input tokens, transforms the thoughts back into embedding space, and adds them to the following input tokens. This has two key advantages. First, it captures the recurrent nature of CoT, but where the thought tokens are generated as input is processing. Second, since the thoughts are represented as tokens, they can be learned from natural language supervision, and using teacher-forcing, which is parallelizable. 
Empirically, Thinking States outperforms other latent reasoning methods on multiple reasoning tasks, narrowing the gap to CoT on math problems, and matching its performance on 2-Hop QA with improved latency. On state-tracking tasks, we show Thinking States leads to stronger reasoning behavior than CoT, successfully extrapolating to longer sequences than seen during training.
\end{abstract}

\section{Introduction}\label{intro}
\input{intro_v2}

\begin{figure*}[t!]
  \begin{center}
    \begin{subfigure}[t]{\columnwidth}
      \centering
      \includegraphics[width=\columnwidth]{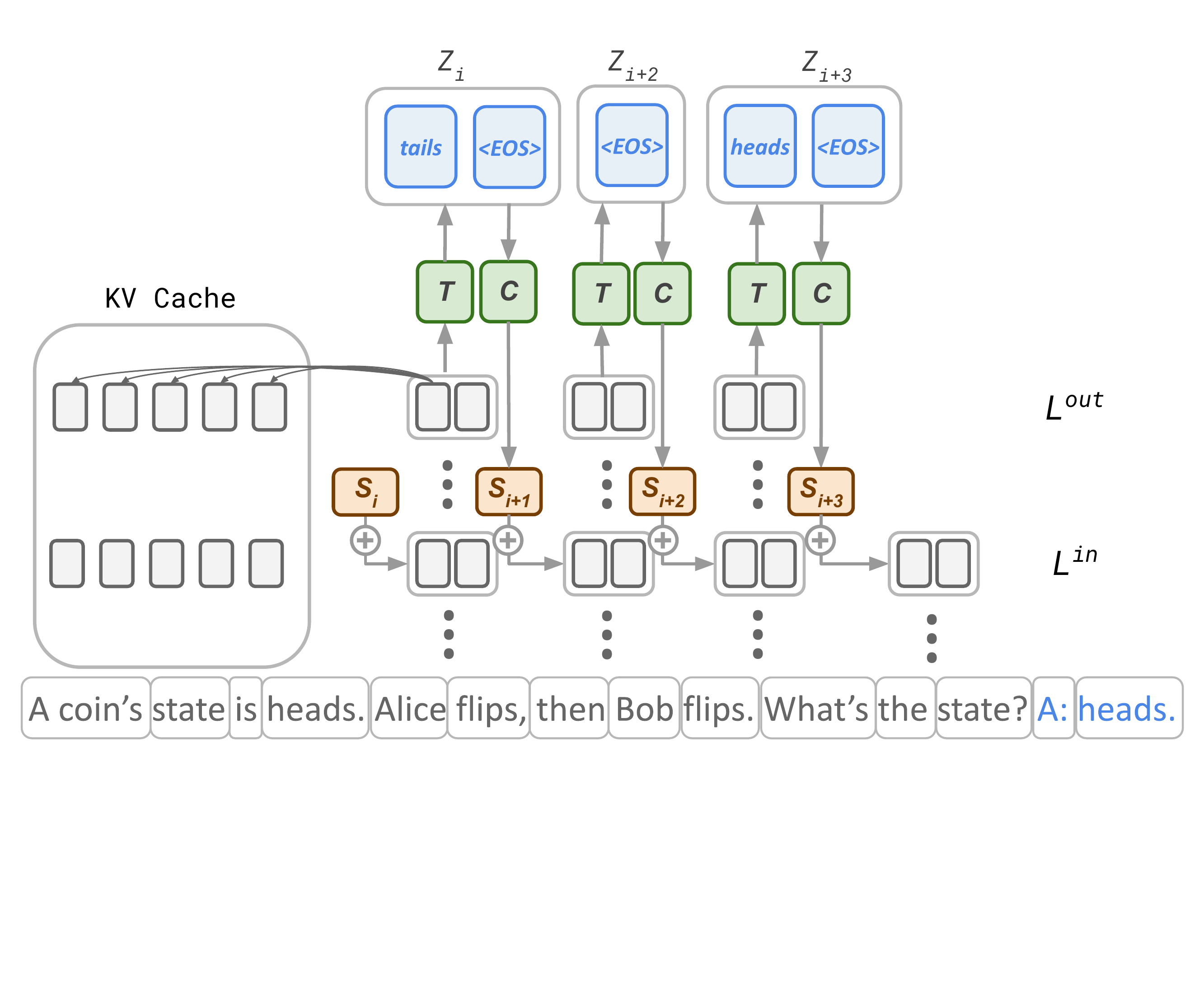}
      \caption{Inference}
      \label{fig:inference-view}
    \end{subfigure}
    \begin{subfigure}[t]{\columnwidth}
      \centering
      \includegraphics[width=\columnwidth]{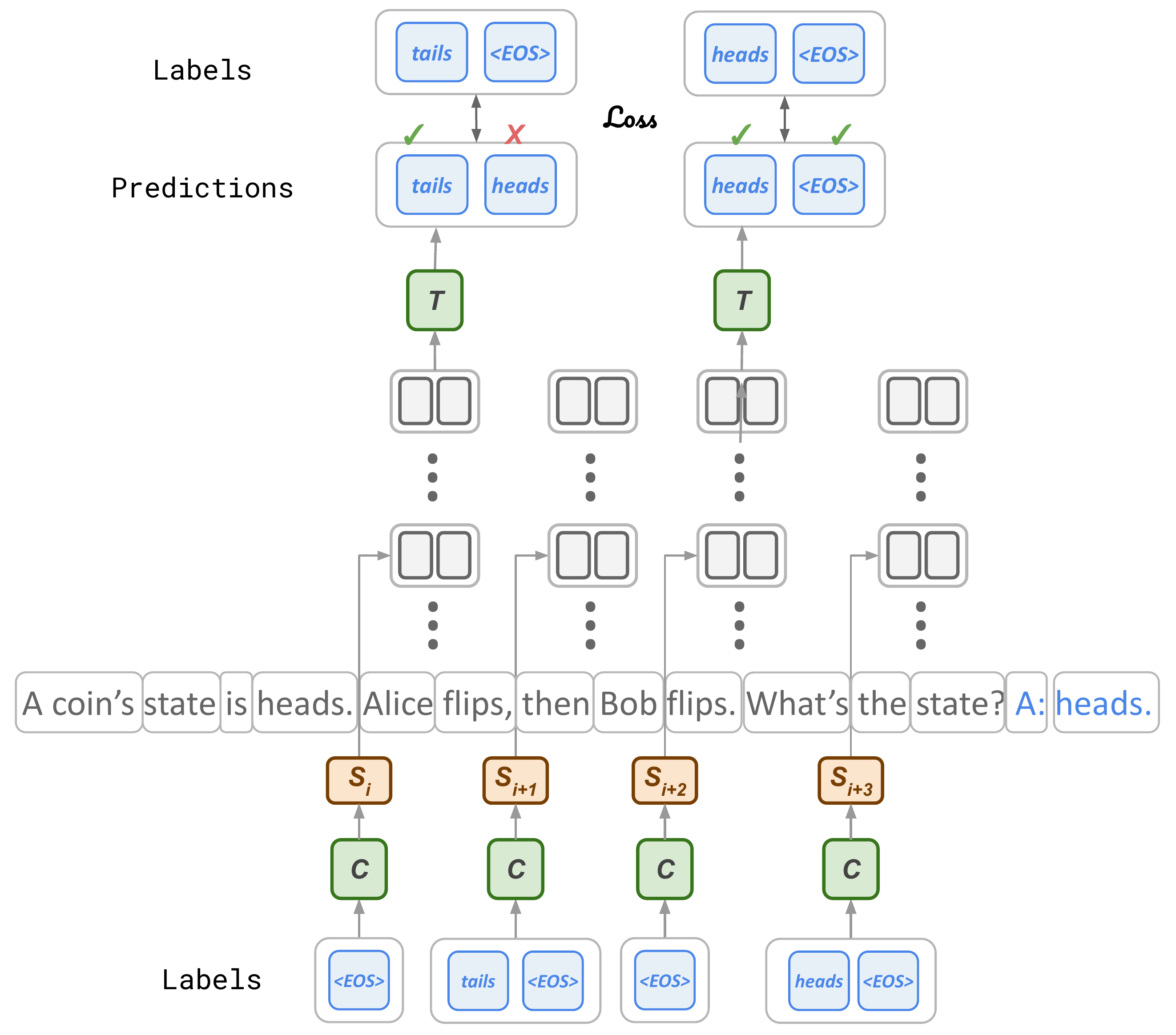}
      \caption{Training}
      \label{fig:training-view}
    \end{subfigure}
    \caption{\textit{\arch} model at inference and training time. At inference time \textit{(a)}, \textit{\arch} reasons by iteratively processing token chunks. At each iteration $i$, reasoning information encoded up to layer $L^{out}$ is decoded by a thinking block {\tb} and compressed by a compression block \cb to a fixed-size \textit{thinking state} $\mathbf{S_{i+1}}$. The state is injected to the next chunk at a shallow layer to effectively influence future computations. Chunk representations can access the history via attention layers and the KV-cache.
    At training time \textit{(b)}, predictions by {\tb} are supervised by explicit annotations paired with each chunk. The same annotations are used to condition the base LLM's reasoning, allowing parallel training.}
  \end{center}
\end{figure*}

\section{Related Work}\label{related-work}
\input{related_work_avi}

\section{Thinking State Architecture and Supervision}\label{method}

\input{method_avi}

\input{data_construct_avi}

\section{Experiments}\label{exps}

\input{experiments}

\section{Conclusions}\label{discuss}
\input{discussion_future}

\bibliography{think_state}
\bibliographystyle{icml2026}

\newpage
\appendix
\input{appendix}
\onecolumn


\end{document}

%% file: intro_v2.tex
Generating intermediate reasoning steps prior to answering enables large language models (LLMs) to tackle challenging problems with considerably greater accuracy \cite{wei2023chainofthoughtpromptingelicitsreasoning}. Such chain-of-thought (CoT) reasoning yields substantial performance gains, but may incur significant inference-time cost because of generating additional tokens. Thus, the challenge of reducing computational cost of reasoning while preserving accuracy has attracted considerable attention recently \cite{xia2025tokenskipcontrollablechainofthoughtcompression, kang2024c3otgeneratingshorterchainofthought, hassid2025dontoverthinkitpreferring, hao2024traininglargelanguagemodels}.

\begin{figure*}[t!]
  \vskip 0.3in
  \begin{center}
    \centerline{\includegraphics[width=\linewidth]{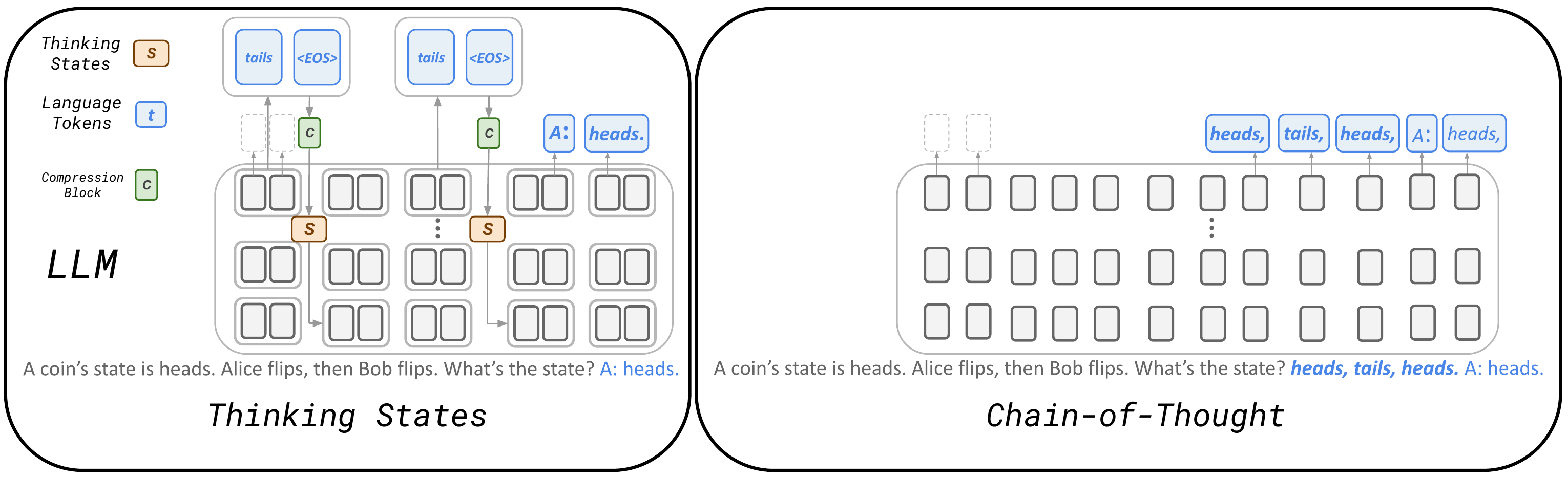}}
    \caption{Reasoning with a \textit{Thinking State} compared to Chain-of-Thought: An LLM is trained to generate task-relevant thinking \emph{sequences} while processing tokens. Each generated thought sequence is transformed into a fixed-size state, denoted by $S$, added to the following token representations.}
    \label{fig:intro}
  \end{center}
  \vskip -0.2in
\end{figure*}

One approach to reducing reasoning tokens is to not require thinking tokens to be in natural language, but rather be continuous embeddings \cite{hao2024traininglargelanguagemodels, shen2025codicompressingchainofthoughtcontinuous}. Indeed, language is highly redundant, and one can therefore expect shorter reasoning traces in this case. 
On the other hand, it is then harder to supervise these ``continuous latent'' thoughts, as target values for the latents are unknown, which requires backpropagation through time (BPTT) from the desired outcome, making training considerably more challenging and limiting the number of latents that can be used in practice.
An alternative approach has been to distill the thinking pattern directly to the representations of the query, rather than generate any extra tokens \cite{deng2024explicitcotimplicitcot}. While supporting parallel training, the resulting architecture cannot effectively condition on thoughts.
As queries are processed in parallel, sequential thinking can only be realized through the LLM's depth, which is typically fixed, resulting in reduced expressivity compared to CoT models.\footnote{Put differently, CoT can be viewed as providing the model with additional depth, which \citet{deng2024explicitcotimplicitcot} does not benefit from.}

Motivated by these previous works we present an architecture that allows conditioning on thinking as in CoT, yet does not append thinking tokens to the context or suffer from optimization with BPTT.
Our \textit{\arch} model builds on several key ideas, illustrated in Figure~\ref{fig:intro}.
{\bf Compute Sharing:} thoughts are generated using representations of existing tokens.
{\bf Recurrence:} these thoughts are fed back as input (as in regular CoT) along with future tokens, not increasing the context length.
{\bf Supervision with Natural Language:} the thought tokens can be supervised via natural language, and thus benefit from available supervision used by prior CoT methods, and maintain interpretability.  

We evaluate {\arch} on synthetic and real-world datasets, demonstrating that it is more accurate than baselines that reduce CoT lengths, and provides a significant latency advantage compared to CoT. Moreover, we demonstrate that {\arch} can be trained significantly faster than models with latent thoughts that require BPTT.

Our contributions are as follows:
\begin{enumerate}[label=(\roman*), itemsep=0pt, topsep=0pt, parsep=0pt]
\item We introduce \arch, a recurrent reasoning mechanism that generates natural-language thoughts while processing the input, without increasing the context length.
\item Because thoughts are represented in natural language, \arch can be trained with teacher-forcing the thoughts, enabling fully parallel training and avoiding the computational overhead of backpropagation through time.
\item We demonstrate that \arch consistently outperforms existing latent reasoning methods on various benchmarks, matches or exceeds CoT accuracy on multi-hop QA and state-tracking tasks, and achieves significant speedups compared to CoT on all tasks.
\end{enumerate}

%% file: related_work_avi.tex
\paragraph{Latent Reasoning.}
A prominent line of research focuses on adapting pretrained LLMs to internalize chain-of-thought (CoT) reasoning into compressed representations.
Initial efforts distilled knowledge from a CoT-capable teacher model directly into the query representations of a student model~\cite{deng2023implicitchainthoughtreasoning}.
A related approach employs an iterative training curriculum that progressively removes explicit reasoning steps from the supervision signal, compelling the model to internalize the reasoning process~\cite{deng2024explicitcotimplicitcot}.
\citet{hao2024traininglargelanguagemodels} extended this idea by introducing recurrent ``continuous thought tokens'' that are iteratively refined after processing the query.
While promising, this recurrent formulation necessitates backpropagation through time (BPTT), which is computationally intensive and limits scalability.
More recently, \citet{shen2025codicompressingchainofthoughtcontinuous} proposed adding a distillation loss that aligns the student's latent representations with those of a CoT-based teacher, removing reliance on the iterative curriculum.
However, this approach still requires BPTT for optimization and incurs the additional computational cost of a separate teacher model.

\paragraph{Emergent Latent Reasoning in Standard LLMs.}
Recent studies have revealed that standard LLMs can exhibit latent reasoning capabilities without explicit training.
\citet{biran2024hoppinglateexploringlimitations} demonstrated this by successfully decoding intermediate reasoning steps from the model's hidden states, while \citet{yang2025largelanguagemodelsperform} showed that LLMs can arrive at correct answers on queries that inherently require multi-hop reasoning.
These findings suggest that some degree of latent reasoning emerges naturally during pretraining, making multi-hop question answering, which we include in our evaluation, an important testbed for methods designed to enhance this capability.

\paragraph{Alternative Approaches to Latent Computation.}
A distinct line of work aims to improve reasoning by providing models with additional computation time, rather than by internalizing a specific reasoning trace.
\citet{goyal2024thinkspeaktraininglanguage} introduced ``pause tokens'' during pretraining---tokens without semantic meaning that give the model extra computational steps before producing output, improving performance on downstream reasoning tasks.
\citet{geiping2025scalingtesttimecomputelatent} proposed dynamic depth, where the same block of layers is applied repeatedly for a variable number of iterations, effectively increasing computation per token at inference time.
Finally, \citet{zelikman2024quietstarlanguagemodelsteach} enabled token-level rationale generation by decoding auxiliary reasoning tokens before each prediction, training the model with reinforcement learning to produce helpful rationales.
Unlike our approach, these auxiliary tokens influence only the current prediction and are discarded afterward, avoiding recurrent dependencies but also forgoing the ability to carry reasoning state across positions.

%% file: method_avi.tex
We introduce the \textit{Thinking States} model, illustrated in Figure~\ref{fig:inference-view}.
The core element of our architecture is a chunk-recurrent process of decoding thoughts from token representations, compressing thoughts into a fixed-size state, and injecting the state into representations of subsequent tokens at a shallow layer (e.g., the first layer).

Because thoughts are decoded from representations in a deep layer (e.g., second to last), we leverage the backbone LLM to perform the bulk of the computation. This allows the decoding \textit{"Thinking Block"} to remain lightweight and efficient, as it operates on a rich feature space.
By injecting the compressed state into representations of future tokens at a shallow layer, we effectively condition on the reasoning process without extending the context.
Sections \ref{method-training}, \ref{method-prefill}, provide details on how this recurrent process is executed efficiently at training and inference time.

\textbf{Setting.}
Let an input sequence of $L$ tokens be partitioned into $K$ non-overlapping chunks $\mathbf{X}_1,\dots,\mathbf{X}_K$.\footnote{In the case $L \bmod c \neq 0$, input tokens are grouped into chunks along with generated tokens, once those are available.} 
Each $\mathbf{X}_i \in \mathbb{R}^{c \times d}$ denotes the input embeddings of $c$ tokens (where $c = L/K$) with hidden dimension $d$, extracted after a designated shallow layer $L^{in}$ of the LLM.

Our architecture augments a backbone LLM $M_\theta$ with two additional modules:

\begin{enumerate}[label=(\roman*), itemsep=2pt, topsep=2pt]
    \item \textit{Thinking Block} {\tb}: A lightweight Transformer decoder (one layer in our experiments) that autoregressively generates a reasoning sequence $\mathbf{Z} = (z_1, \dots, z_n)$ in natural language.
    \item \textit{Compression Block} {\cb}: A Transformer encoder with a pooling layer that maps any variable-length reasoning sequence into a fixed-size state $\mathbf{S} \in \mathbb{R}^{c \times d}$.
\end{enumerate}

Further architectural details for {\tb} and {\cb} are provided in Appendix~\ref{app-arch}.

\subsection{Architecture}\label{method-arch}

\arch processes token chunks iteratively while maintaining a state $\mathbf{S}_i \in \mathbb{R}^{c \times d}$.
At each step $i$, we first inject the current state into the input representations at the shallow layer, then forward through $M_\theta$:
\begin{equation}\label{eq-step-update}
    \tilde{\mathbf{X}}_i = \mathbf{X}_i + \mathbf{S}_i
\end{equation}
\begin{equation}\label{eq-pred-hs}
    \mathbf{H}_i^{out}
    = M_\theta(\tilde{\mathbf{X}}_i \mid \tilde{\mathbf{X}}_{<i})
\end{equation}

Here, $\mathbf{H}_i^{out}$ denotes the chunk representations at a predefined deep layer $L^{out}$, and past representations $\tilde{\mathbf{X}}_{<i}$ are accessed through the KV-cache.

Next, $\mathbf{H}_i^{out}$ is used to generate a sequence of natural-language thought tokens via the Thinking Block:
\begin{equation}\label{eq-gen-t}
    \mathbf{Z}_{i+1} = {\tbm}(\mathbf{H}_i^{out})
\end{equation}
The output $\mathbf{Z}_{i+1}$ is a variable-length token sequence ending with an \texttt{<EOS>} token.
When no reasoning is required, $\mathbf{Z}_{i+1}$ may consist solely of \texttt{<EOS>}.

Finally, the Compression Block transforms the variable-length thought sequence into a fixed-size state for the next iteration:
\begin{equation}\label{eq-comp-t}
    \mathbf{S}_{i+1} = {\cbm}(\mathbf{Z}_{i+1}) \in \mathbb{R}^{c \times d}
\end{equation}
For the first chunk, we initialize $\mathbf{S}_1 = \mathbf{0}$.

This design serves two key ideas: compute sharing and recurrence.
As inputs are a sum of tokens and states, eq.~(\ref{eq-step-update}-\ref{eq-pred-hs}), the model effectively performs next token prediction while processing and generating states, executing both computations in a single forward pass.
By injecting states at a shallow layer, we allow thoughts to effectively condition future thoughts through most of the LLM's hidden layers.
Importantly, since the thought tokens are never appended to the backbone's context window, the context length remains fixed.

\subsection{Training with Teacher-Forced Reasoning}\label{method-training}

To supervise the generated thinking sequences $\mathbf{Z}_{i}$, we rely on explicit natural-language annotations.
In our formulation, each token chunk in the training data is paired with a target reasoning sequence $\mathbf{Z}_{i}^*$ of variable length.
For chunks that require no intermediate reasoning, the target is simply the \texttt{<EOS>} token.

Access to these ground-truth sequences $\mathbf{Z}_{i}^*$ serves a dual purpose.
As illustrated in Figure~\ref{fig:training-view}, in addition to providing supervision for the outputs of {\tb}, we use $\mathbf{Z}_{i}^*$ to \emph{teacher-force} the states injected into the backbone LLM.
Specifically, we compute the target state as $\mathbf{S}_{i}^* = {\cbm}(\mathbf{Z}_i^*)$ and inject it instead of the model-generated state:
\begin{equation}
    \tilde{\mathbf{X}}_i = \mathbf{X}_i + \mathbf{S}_i^*
\end{equation}

Since all target sequences $\mathbf{Z}_{i}^*$ are available upfront, we can compute all chunk representations $\mathbf{H}_i^{out}$ in a single parallel forward pass through $M_\theta$.
This eliminates the need for backpropagation through time (BPTT), dramatically reducing the computational cost of training.

After this forward pass, the Thinking Block {\tb} is trained to predict $\mathbf{Z}_{i}^*$ via standard next-token prediction, in parallel over all chunks.
Here, each $\mathbf{H}_i^{out}$ is computed under the gold state history $\mathbf{S}_{1}^*, \dots, \mathbf{S}_{i}^*$, so predicting $\mathbf{Z}_{i+1}$ is implicitly conditioned on all prior gold reasoning steps.

We optimize the joint objective:
\begin{equation}
    \mathcal{L} = \mathcal{L}_{\text{LM}} + \sum_{i=1}^{K} \mathcal{L}_{\text{T}}(\mathbf{Z}_{i}, \mathbf{Z}^*_{i})
\end{equation}
where $\mathcal{L}_{\text{LM}}$ is the standard language modeling loss and $\mathcal{L}_{\text{T}}$ is cross-entropy loss over the thinking sequences.
This formulation enables fully parallel training while ensuring that {\tb} learns to generate thoughts consistent with a valid reasoning history.

\subsection{Fast Prefill with Speculative Thinking}\label{method-prefill}

While training is fully parallelizable, na\"ive inference remains sequential across chunks, as each $\tilde{\mathbf{X}}_i$ depends on the previous state $\mathbf{S}_{i}$.
This can significantly increase latency during query prefill.

To mitigate this overhead, we exploit a key observation: for most chunks, the generated state is \emph{trivial}, that is, the Thinking Block produces only an \texttt{<EOS>} token.
We leverage this sparsity to design an iterative yet \textit{exact} prefill algorithm.

The algorithm proceeds as follows:
\begin{enumerate}[itemsep=2pt, topsep=2pt]
    \item Perform a parallel forward pass over all chunks, speculating that all states are trivial.
    \item For each chunk, generate thinking states using {\tb} and {\cb}.
    \item Identify the earliest chunk $i_1$ that produces a non-trivial state. Since all chunks before $i_1$ truly have trivial states, the computation up to chunk $i_1$ is correct.
    \item Cache all positions up to $i_1$ and repeat from step 1 for the remaining chunks, now conditioning on the correct state $\mathbf{S}_{i_1}$ and speculating a trivial state for following chunks.
\end{enumerate}

The procedure terminates when no new non-trivial states are generated, yielding exactly the same result as full sequential prefill.
If $|R|$ denotes the (unknown a priori) number of chunks with non-trivial states, the algorithm completes in exactly $|R|+1$ rounds.
In typical regimes where $|R| \ll K$, this substantially reduces prefill latency.
A complete description of the algorithm is provided in Appendix~\ref{app-fast-prefill}.

%% file: data_construct_avi.tex
\subsection{Constructing Chunk-Level Supervision}\label{sec-data-gen}

Training the Thinking Block requires supervision that specifies both \emph{where} intermediate reasoning should occur and \emph{what} reasoning content to produce.
We construct this supervision by aligning CoT reasoning steps to specific positions in the input query.

Consider the following example query and its CoT trajectory:
\begin{quote}
\textit{``A coin's state is heads. Alice flips, then Bob flips. What's the state?''}\\[2pt]
CoT: \textit{``heads $\rightarrow$ tails $\rightarrow$ heads''}
\end{quote}
Our goal is to map each reasoning step (which may span multiple tokens) to a specific position in the query where that step becomes inferable.

\paragraph{Step-to-Token Alignment.}
We first align each reasoning step to the earliest query position where it can be logically inferred.
This alignment is obtained either via a strong teacher model, e.g., Gemini 2.5-Flash \cite{comanici2025gemini}, or through programmatic rules when the task structure permits.
The alignment is represented by inserting special indicator tokens \textit{\texttt{<T>}} into the text:
\begin{quote}
\textit{``A coin's state is heads.\texttt{<T>} Alice flips,\texttt{<T>} then Bob flips.\texttt{<T>} What's the state?''}
\end{quote}
Each \textit{\texttt{<T>}} marker indicates that the preceding context is sufficient to infer the corresponding reasoning step.
We then tokenize the query and construct a corresponding reasoning array with the same length, where each position in the array is either empty or contains the reasoning step associated with the \textit{\texttt{<T>}} token.
The indicator tokens are removed after this alignment, ensuring they are never seen by the model during training.

\paragraph{Token-to-Chunk Alignment.}
Given the aligned token sequence and reasoning array, we partition both into non-overlapping chunks of size $c$.
The target reasoning sequence for each chunk is the concatenation of all reasoning steps assigned to tokens within that chunk.
For chunks with no associated reasoning steps, the target is simply the \texttt{<EOS>} token.

The entire process is applied only to the training data of each task we consider.
Additional details can be found in Appendix~\ref{app:data-gen}.

%% file: experiments.tex
We evaluate \arch on a diverse set of tasks that probe both reasoning accuracy and computational efficiency. Section~\ref{res-state} examines state-tracking, where \arch exhibits stronger length generalization than CoT. Section~\ref{res-reasoning} evaluates general reasoning on Multi-Hop QA and GSM8K, showing that \arch surpasses latent reasoning baselines and approaches CoT performance with significant speedups. Section~\ref{res-ablation-studies} presents ablation studies exploring the performance--efficiency tradeoff as a function of recurrence depth and chunk size. Finally, Section~\ref{res-error-anal} provides an interpretability-driven error analysis, identifying distinct failure modes and cases where \arch outperforms CoT.

\subsection{Experimental Setup}\label{sec-exp-setup}

All experiments use \textit{Qwen2.5-Base} models~\cite{qwen2025qwen25technicalreport} (0.5B and 1.5B parameters), fine-tuned on task-specific training data.

We compare against four baselines:
\begin{enumerate}[label=(\roman*), itemsep=1pt, topsep=2pt, parsep=0pt]
    \item \textbf{CoT}: fine-tuning on explicit chain-of-thought traces;
    \item \textbf{No CoT}: standard fine-tuning to predict the final answer directly;
    \item \textbf{iCoT}~\cite{deng2024explicitcotimplicitcot}: a latent reasoning method that distills CoT into query representations via curriculum learning;
    \item \textbf{Coconut}~\cite{hao2024traininglargelanguagemodels}: a state-of-the-art latent reasoning method using continuous thought tokens.
\end{enumerate}
We reproduce iCoT and Coconut results using the official codebase of~\citet{hao2024traininglargelanguagemodels} with default configurations unless otherwise noted.

\paragraph{Metrics.}
We report accuracy on gold answers across all tasks.
For efficiency, we measure wall-clock speedup relative to CoT on a single A100-80GB GPU.
Unlike prior work~\cite{hao2024traininglargelanguagemodels, deng2024explicitcotimplicitcot}, which reports efficiency in terms of generated token count, we use wall-time because token generation with the lightweight Thinking Block {\tb} is substantially faster than standard autoregressive decoding.

\input{exp_state_tracking_exps}

\input{exp_general_reasoning_exps}


\input{exp_ablations}

\input{exp_error_analysis}

%% file: exp_state_tracking_exps.tex
\subsection{State Tracking Tasks}\label{res-state}
State tracking tasks require monitoring quantities that evolve throughout the input~\cite{illusionofstate, kim2023entitytrackinglanguagemodels, li2025howlanguagemodelstrack}.
These tasks are a natural testbed for latent reasoning methods because they explicitly require \emph{sequential composition}---each step depends on the result of previous steps.
We evaluate on two synthetic tasks adapted from~\citet{anil2022exploringlengthgeneralizationlarge}:
\begin{itemize}[itemsep=1pt, topsep=2pt, parsep=0pt]
    \item \textit{Parity}: tracking a single binary state through a sequence of flip operations.
    \item \textit{Variable Assignment (Vars)}: tracking multiple interacting integer variables, yielding more complex state dynamics.
\end{itemize}
Both tasks are expressed in natural language; formal definitions and examples are provided in Appendix~\ref{app-state-track}.
\begin{table}[t]
  \centering
  \caption{Performance (accuracy) on State Tracking tasks, trained on samples with up to $N$ state updates and evaluated on $[N, 100]$.}
  \resizebox{\columnwidth}{!}{%
    \begin{tabular}{@{}lcccccc@{}}
      \toprule
      & \multicolumn{3}{c}{Parity} & \multicolumn{3}{c}{Vars}\\
      \cmidrule(lr){2-4} \cmidrule(lr){4-6}
      Method        & N=10  & N=20  & N=40  & N=10  & N=20  & N=40 \\
      \midrule
      No CoT        & 54.67     & 57.50     & 59.60         & 02.15    & 02.17    & 02.19 \\
      CoT           & 12.35         & 38.12         & 64.38         & 06.78     & 35.45     & 87.75 \\
      \arch        & 98.37    & 99.02    & 100.00    & 33.76     & 87.23    & 97.71 \\
      \bottomrule
    \end{tabular}%
  }
    \label{state-tracking-results}
\end{table}

To assess whether the underlying base model truly learns to reason via the recurrent thinking state mechanism, we employ a length generalization setting, drawing on the known extrapolation capabilities of recurrent models~\cite{delétang2023neuralnetworkschomskyhierarchy}. We fine-tune all methods, with \textit{Qwen2.5-Base-0.5B} as the base model, on sequences of up to $N$ operations ($N \in \{10, 20, 40\}$) and evaluate on lengths in $[N, 100]$, in both cases, sampled uniformly over the range. All models are trained to $100\%$ in-distribution accuracy to strictly isolate extrapolation capabilities from optimization effects.

Table~\ref{state-tracking-results} reports out-of-distribution (OOD) accuracy for each training regime.
The results demonstrate how a base LLM trained with \arch effectively adapts to reasoning with a recurrent mechanism, leading to stronger length generalization than that of a model reasoning via CoT.
Critically, this indicates that reasoning via the recurrent mechanism offers a more effective solution than CoT, requiring a simpler training distribution to achieve robust performance.

Furthermore, while CoT generates intermediate tokens proportional to the number of state operations, considerably extending the context length and thus expressivity, \arch maintains a static context size. By refining the query representation through state updates, \arch achieves superior length generalization, suggesting the model has learned to reason through the recurrent mechanism.

%% file: exp_general_reasoning_exps.tex
\subsection{General Reasoning Capabilities}\label{res-reasoning}

\begin{table}[t]
  \centering
  \caption{Performance and speedup (over CoT) on reasoning tasks for different latent reasoning baselines and CoT.}
  \setlength{\tabcolsep}{2.2pt}
  \resizebox{\columnwidth}{!}{%
    \begin{tabular}{@{}lcccccc@{}}
      \toprule
      & \multicolumn{2}{c}{GSM} & \multicolumn{2}{c}{2-Hop \textit{FK}} & \multicolumn{2}{c}{2-Hop \textit{PK}} \\
      \cmidrule(lr){2-3} \cmidrule(lr){4-5} \cmidrule(lr){6-7}
      Method & Acc$\uparrow$ & Speedup$\uparrow$ & Acc & Speedup & Acc & Speedup \\
      \midrule
      CoT                         & 60.50    & $\times$1           & 54.79            & $\times$1       & 43.07         & $\times$1     \\
      \midrule
      No CoT                      & 34.11    & $\times$5.59        & 33.47            & $\times$1.89    & 31.92         & $\times$2.03  \\
      \arch                       & 42.22    & $\times$2.66        & 54.91            & $\times$1.19    & 43.05         & $\times$1.23  \\
      Coconut                     & 32.65    & $\times$3.14        & 33.71            & $\times$1.14    & 32.60         & $\times$1.21  \\
      iCoT                        & 34.00    & $\times$5.71        & 28.84            & $\times$1.59    & 36.31         & $\times$1.80  \\
      \bottomrule
    \end{tabular}%
  }
  \label{reasoning-results}
\end{table}

We evaluate general reasoning capabilities on two established benchmarks.
First, a 2-hop QA dataset, where prior work by \citet{biran2024hoppinglateexploringlimitations, yang2025largelanguagemodelsperform} demonstrated standard LLMs exhibit latent reasoning to some extent, providing a useful test case for methods designed to improve latent reasoning.
Second, a GSM8K-style dataset, consisting of $\sim$400K simple math word problems and parsed CoT steps, previously used by ~\citet{deng2023implicitchainthoughtreasoning, hao2024traininglargelanguagemodels}.
Methods in this section use \textit{Qwen2.5-Base-1.5B} as the base model.

For the 2-hop QA task, we consider two evaluation regimes. 
In \textit{Full Knowledge (FK)}, the required factual knowledge appears in the fine-tuning data, testing how well models acquire and manipulate knowledge with each method. 
In \textit{Parametric Knowledge (PK)}, examples are filtered to reflect the knowledge of the base LLM, isolating whether a method improves retrieval and manipulation of existing knowledge  in the base model.

\begin{figure*}[t!]
  \begin{center}
    \begin{subfigure}[t]{\columnwidth}
      \centering
      \includegraphics[width=0.9\columnwidth]{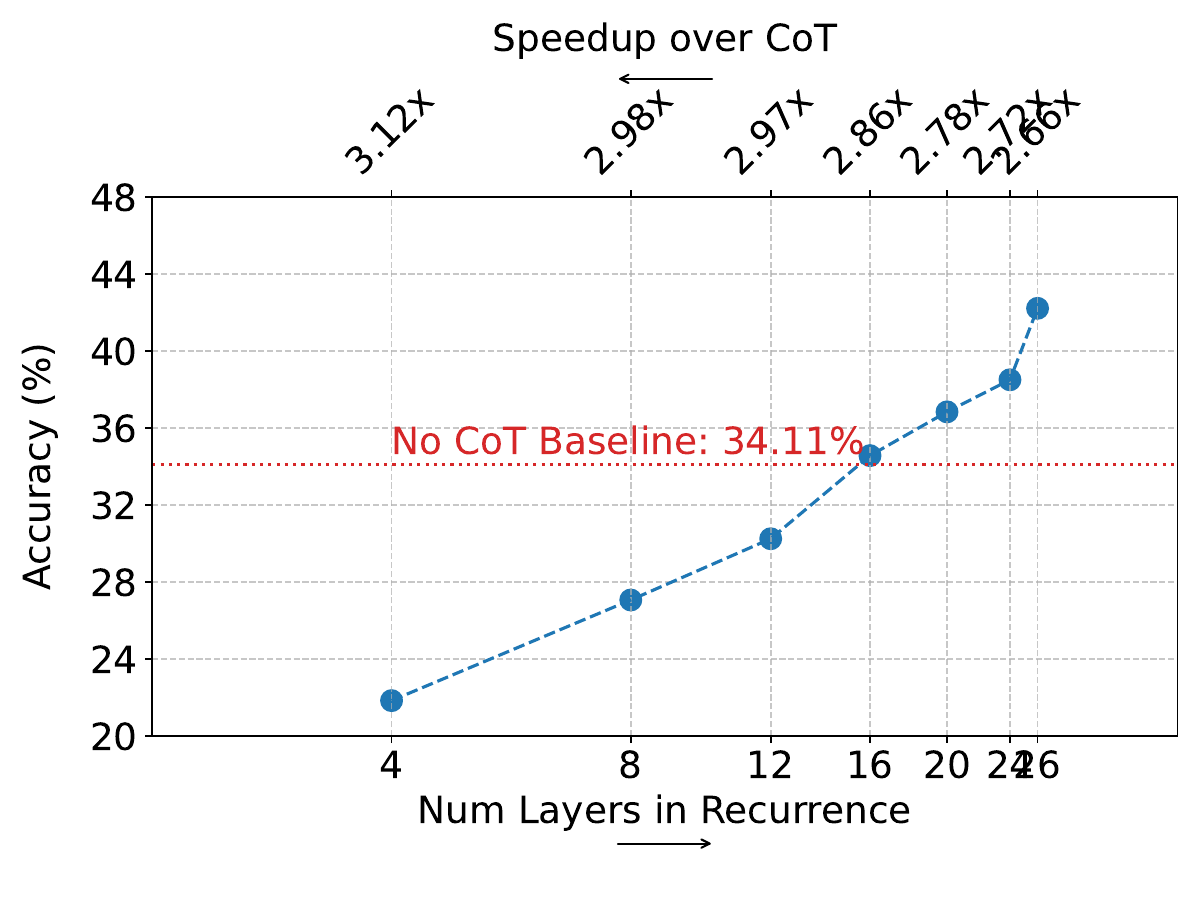}
      \caption{}
      \label{fig:layer-ablat}
    \end{subfigure}
    \begin{subfigure}[t]{\columnwidth}
      \centering
      \includegraphics[width=0.9\columnwidth]{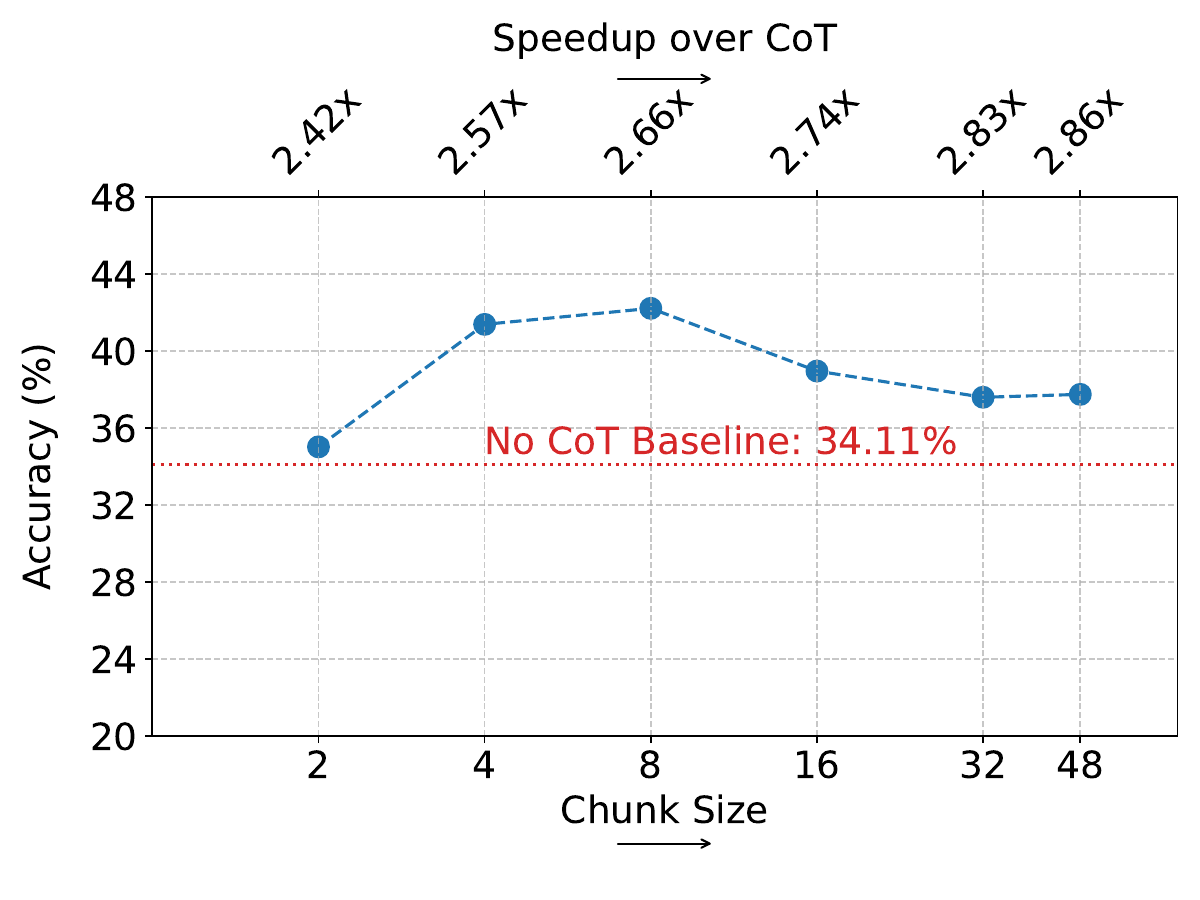}
      \caption{}
      \label{fig:chunk-ablat}
    \end{subfigure}
    \caption{Ablation studies over deep to shallow recurrence \textit{(a)} and effects of the chunk size \textit{(b)}, with measured speedup over CoT (top) at each point. In \textit{(a)}, performance increases with the number of layers used to process and generate states, pointing to the importance of the deep to shallow design. In \textit{(b)}, increasing the chunk size leads to improved latency with peak performance for medium chunks, highlighting the tradeoff between per-state capacity (large chunks) and reasoning frequency (small chunks).}
  \end{center}
\end{figure*}

Table~\ref{reasoning-results} summarizes accuracy and speedups for each method.
Among latent-reasoning baselines, \arch consistently achieves the highest accuracy across all tasks while maintaining significant speedups compared to CoT.
On both variants of the 2-Hop task, \arch reaches similar performance to CoT with improved efficiency, demonstrating a dramatic advantage over other non-CoT baselines, with more than a 20\% and 7\% increase in accuracy on the \textit{FK} and \textit{PK} variants respectively.

On the GSM task, \arch demonstrates considerably better performance compared to Coconut, No CoT and iCoT, improving accuracy by $\sim8\%$ and maintaining a considerable $2.66\times$ speedup.
The slight latency advantage showcased by Coconut on this task leads us to evaluate if performance can be improved at the cost of higher latency.
As reported in Appendix~\ref{app-coco-res}, we observe that increasing the number of latent steps leads to a modest improvement in performance, even as latency approaches that of the CoT model.
In the next section, we provide several ablation studies over possible compute-efficiency tradeoffs with \arch.

Finally, we analyze the training costs of \arch compared to training with explicit recurrence and BPTT as in Coconut. Figure~\ref{fig:training-compute}, in Appendix~\ref{app-training-efficiency}, plots the wall-clock time per forward and backward pass as a function of the number of recurrent steps/number of chunks in the input sequence. The results demonstrate that while BPTT incurs linearly increasing costs, \arch maintains close to constant training time due to teacher forcing. Consequently, while BPTT becomes prohibitively expensive to scale, incurring a $\sim10\times$ cost penalty at just 10 steps, \arch enables efficient training regardless of reasoning depth.


%% file: exp_ablations.tex
\subsection{Ablation Study}\label{res-ablation-studies}
To understand the architectural factors governing \arch, we conduct two ablation studies over the GSM task.
We ablate the impact of recurrence with a deep to shallow connection, by varying the layer to extract representations passed to \tb, and the impact of reasoning frequency, by varying the chunk size.

\textbf{Impact of Deep to Shallow Connections:} In Figure~\ref{fig:layer-ablat}, we investigate the influence of deep to shallow recurrence by varying the extraction layer  for \tb, keeping injection fixed at the first layer (as in previous experiments).
Our purpose is to show the importance of the deep to shallow recurrent principle, demonstrating that allocating more of the model's capacity when processing and generating states leads to improved accuracy.
Furthermore, this experiment highlights the inherent accuracy-latency tradeoff: utilizing fewer layers in the recurrent loop allows the remaining layers to be executed only during query prefill, resulting in higher speedups. For this analysis, we use \textit{Qwen2.5-Base-1.5B}, which contains 28 hidden layers.

As shown in Figure~\ref{fig:layer-ablat}, performance increases monotonically with the number of layers included in the recurrence, with a substantial accuracy gap of nearly 20\% between the shallowest and deepest configurations. This trend confirms that maximizing the computational capacity available for processing and generating thinking states is critical for performance, effectively validating the deep-to-shallow design choice.

\textbf{Impact of Chunk Size:} In Figure~\ref{fig:chunk-ablat}, we study the effect of varying the chunk size.
The chunk size controls two important factors: \textit{(1)} the computational capacity available for state processing and generation \textit{(2)} the number of \textit{expected} reasoning steps generated in each call to \tb.
A small chunk size limits available capacity for each state, and is expected to generate less steps at each chunk.
Conversely, larger chunks provide larger capacity, but may lead to encoding multiple sequential steps into single state.
This latter scenario undermines the deep-to-shallow recurrence, as consecutive steps generated within the same chunk cannot access the full recurrent loop, relying instead on the LLMs depth and limited capacity of the lightweight Thinking Block.

The tradeoff discussed above is evident in Figure~\ref{fig:chunk-ablat}, when the chunk size is small both performance and latency degrade, which can be explained by lack of computational capacity and increased iterations during query processing.
As the chunk size increases, latency improves due to fewer iterations during query processing.
Increasing the chunk size beyond a certain point ($c=8$) eventually leads to reduced performance, which can be explained by the model compressing too many reasoning steps into single updates.
Although performance decreases, we note that \arch still maintains a notable gap in performance compared to other latent reasoning methods even when approaching similar speedups.

%% file: exp_error_analysis.tex
\subsection{Error Analysis}\label{res-error-anal}



We analyze the performance gap between \arch and CoT on GSM-style problems.
A key advantage of \arch over other latent reasoning methods is interpretability: because states are compressed from natural-language thoughts, the model's reasoning remains inspectable.
\paragraph{Where \arch succeeds and CoT fails.}
Despite the overall accuracy gap, approximately 12\% of queries correctly solved by \arch are \emph{not} solved by CoT, indicating that \arch mitigates certain CoT errors.
Figure~\ref{fig:cot_fails} shows two representative examples:
(1) CoT hallucinates an additional reasoning step, and
(2) CoT attempts an overly complex intermediate computation.
In both cases, \arch produces a correct internal trajectory and answer.
We note that both models are fine-tuned from the same base model and on identical data.

\begin{figure}[t]
  \centering
  \small
  \setlength{\fboxsep}{6pt}
  \setlength{\fboxrule}{0.6pt}

  \fbox{%
    \begin{minipage}[t]{\dimexpr\columnwidth-3\fboxsep-2\fboxrule\relax}
    \ttfamily
    \raggedright

    \textbf{Example 1: Hallucinated Step}\par
    Derrick has a bakery that makes 10 dozen doughnuts every day,\\
    \textcolor{green!50!black}{[T: 10*12=120]} selling at \$2 per doughnut.\textcolor{green!50!black}{[T: 120*2=240]} How much money does Derrick make in June if he sells all the doughnuts?\textcolor{green!50!black}{[T: 240*30=7200]}\textcolor{blue}{ The answer is 7200}\\
    \textbf{CoT Steps:} 10*12=120 120*2=240 240*30=7200 \textbf{7200*6=43200}
    \vspace{6pt}

    \textbf{Example 2: Over-Complex Computation}\par
    Jess is trying to guess the number of blue jellybeans in a jar. She can see that there are 17 green jelly beans and twice as many red jelly beans.\\
    The\textcolor{green!50!black}{[T: 17*2=34]} rest of the jellybeans are blue jelly beans. If there are a total of 60 jelly beans in total,\\
    \textcolor{green!50!black}{[T: 17+34=51]} how many blue jellybeans are there?\textcolor{green!50!black}{[T: 60-51=9]}\textcolor{blue}{ The answer is 9} \\
    \textbf{CoT Steps:} 17*2=34 \textbf{60-17-34=8}
    \end{minipage}
  }

  \caption{Examples where \arch succeeds but CoT fails. \arch reasoning is shown in \textcolor{green!50!black}{green}. (1) CoT hallucinates an extra step. (2) CoT attempts multiple operations in one step and errs.}
  \label{fig:cot_fails}
\end{figure}


\newpage
\paragraph{Where CoT succeeds and \arch fails: State Ambiguity.}
We identify a distinct error pattern we term \emph{state ambiguity}: in many GSM queries, the quantity of interest is specified only in the final clause and may be difficult to anticipate earlier.
As illustrated in Figure~\ref{fig:state_ambiguity_example}, this leads \arch to reason about the wrong intermediate quantity.
Although the correct answer is often recoverable via a simple transformation, such recovery requires exposure to similar patterns during training.







\begin{figure}[t]
  \centering
  \small
  \setlength{\fboxsep}{6pt}
  \setlength{\fboxrule}{0.6pt}

  \fbox{%
    \begin{minipage}[t]{\dimexpr\columnwidth-3\fboxsep-2\fboxrule\relax}
    \small
    \ttfamily\raggedright

    \textbf{(a) Original}\par
    Richard lives in a building with 15 floors. Each floor contains 8 units.\\
    \textcolor{green!50!black}{[T: 15$\times$8=120.]}Three quarters of the units are occupied.
    \textcolor{green!50!black}{[T: (3/4)$\times$120=90.]} What's the number of unoccupied units on each floor?\textcolor{green!50!black}{[T: 120-90=30.]} \textcolor{blue}{Answer:30}

    \vspace{6pt}

    \textbf{(b) Disambiguated}\par
    \textit{What's the number of unoccupied units on each floor?} Richard lives in a building with 15 floors. Each floor contains 8 units. 
    Three quarters of the units are occupied.\\
    \textcolor{green!50!black}{[T: 0.75$\times$8=6]}What's the number of unoccupied units on each floor? \textcolor{blue}{Answer:2}

    \end{minipage}
  }

  \caption{Illustration of \emph{state ambiguity}. In \text{(a)}, \arch generates valid thoughts that are not aligned with the final clause in the query, leading to an error. By adding the information to the begining of the query, as in \textit{(b)}, the same model produces a correct answer, without training on the new format.}
  \label{fig:state_ambiguity_example}
\end{figure}

We validate this hypothesis by prepending the final question to the start of the prompt for all error cases. This zero-shot intervention improves accuracy from 42.22 to 48.65, despite the augmented queries being out-of-distribution. Notably, this limitation stems from the combination of \arch with a causal autoregressive backbone rather than the thinking-state mechanism itself; bidirectional query processing could, in principle, identify the relevant quantity before committing to intermediate state updates.

%% file: discussion_future.tex
In this work, we introduce Thinking States, a method that enables reasoning while processing input tokens without extending the model's context. By relying on a recurrent state representation of the thinking trajectory, the model effectively conditions its token processing on intermediate reasoning steps  Empirically, this leads to improved performance compared to other latent reasoning baselines while maintaining a significant efficiency advantage over Chain-of-Thought (CoT).

Furthermore, a key contribution of our approach is the specialized data construction process that maps reasoning steps to specific input chunks. This formulation ensures that the recurrent process does not optimize with BPTT during training. Instead, training is fully parallelizable via teacher forcing, leading to a substantial reduction in computational cost compared to prior recurrent methods.

Our framework opens several avenues for future research. First, while we currently target query prefill, extending Thinking States to the decoding phase could enable allocating dynamic compute during token generation. Second, while our current approach relies on Supervised Fine-Tuning over existing CoT data, this model can serve as an effective "warm-start" initialization for Reinforcement Learning. Starting from a model that has already learned to compress reasoning into latent states could stabilize RL training, allowing the model to subsequently optimize its internal thinking process beyond the constraints of human-generated traces.


\section*{Impact Statement}
This paper presents a method for improving the efficiency of reasoning in large language models. By reducing computational costs relative to chain-of-thought approaches, our work may contribute to lowering the energy consumption and environmental footprint of deploying reasoning-capable language models. Additionally, because our method generates interpretable natural-language thoughts, it may support efforts toward more transparent and auditable AI systems. We do not foresee unique ethical concerns beyond those broadly associated with advances in language model capabilities.

%% file: appendix.tex
\section{Method - Additional Details}\label{app-method}

\subsection{Architecture}\label{app-arch}
We provide additional details on the thinking and compression blocks, \tb and \cb and their integration with the base LLM, \llm.
In all reported  experiments, except when explicitly stated in section~\ref{res-ablation-studies}, we set the injection and extraction layer, $L^{in}, L^{out}$, to be after the first layer (excluding the token embedding layer) and after the second-to-last layer respectively.

Both \tb and \cb share the same underlying core module, a causal Transformer block, sharing configurations and hyper-parameters (i.e., hidden size, number of attention heads, sliding/dense attention etc.) as the layers in \llm.
In our experiments, \llm follows the architecture of \textit{Qwen2.5} models, with model sizes listed in the main text.
As the thinking head, \tb, is used for auto-regressive generation of language tokens, and is expected to return token embeddings, it additionally includes an unembedding and embedding layer.

This architecture facilitates the function of the thinking head and compression head as follows.
At inference time, \tb produces thoughts in natural-language through autoregressive generation, initially conditioned on the chunk of hidden representations $\mathbf{H}_i^{out} \in \mathbb{R}^{c \times d}$, where $c, d$ denote the chunk size and the hidden size of \llm.
Each token predicted by \tb is passed through the embedding layer, and the resulting embedding is appended to the context of \tb.
This results in a distribution mismatch, as $\mathbf{H}_i^{out}$ (extracted from a deep layer) share the context with standard token embeddings. However, we observe that this does not lead to a significant limitations when the block is finetuned for next-token-prediction in the same setup, even with relatively little data as in 2-Hop \textit{PK}

For \cb, to use the causal Transformer as a compression module, we apply a single parallel forward pass to all input token embeddings using a causal attention mask (though bidirectional attention is also valid) to obtain a contextualize representations.
We then extract the last $c$ representations to serve as the state $\mathbf{S}_i$ injected to the next chunk, appending padding tokens to the input sequence if it is shorter than $c$.
Through finetuning, \cb learns to compress relevant information into future positions.
We also experimented with appending a set of learnable queries appended to the query to make the compression explicit, however, this approach performed worse in practice.

To initialize the parameters of the two modules, we utilize existing components from \llm. 
Namely, the Transformer block in \tb is initialized as an independent copy of the \emph{last} layer of \llm, while \cb is initialized as an independent copy of the \emph{first} layer (excluding the token embedding layer). 
The unembedding layer in \tb is initialized as an independent copy  of the unembedding layer of \llm while the embedding layer is shared with \llm (i.e., is not an independent copy).

For \tb, the motivation is to leverage parameters already trained to convert hidden representations into token predictions.
For \cb, the motivation is to leverage parameters trained to contextualize token embeddings \cite{kaplan2025tokenswordsinnerlexicon} and generate a state sharing a latent space with the injection layer, which immediately follows the first layer.

\subsection{Fast Prefill with Speculated Thinking}\label{app-fast-prefill}

We provide a more detailed description of the fast prefill algorithm detailed in Section~\ref{method-prefill} in algorithm~\ref{alg:speculative_prefill}\footnote{Full version can be reviewed in our codebase, which we plan to fully release}.
The idea is using the sparsity of non-trivial states, generated by more than an \texttt{<EOS>}, to a few allow parallel execution instead of iteratively processing all chunks in the query.
This is achieved by always speculating future states are trivial, applying a parallel pass conditioned on the speculation, then correcting for the first location where the assumption breaks and repeating.

This process is guaranteed to terminate in exactly $|R| + 1$ iterations, where $|R|$ is the number of non-trivial states generated, which is unknown in advance.
As each iteration after the first implies exactly one non-trivial state was generated, this leads to exactly $|R| + 1$ rounds.

In algorithm \ref{alg:speculative_prefill}, we explicitly denote that \llm returns a set of hidden representations $H$ from all layers, only then extracting the target representation $\mathbf{H}_i^{out}$ to emphasize all representations are stored in the cache.
Additionally, for ease of presentation, the algorithm omits the fact that when finished, the representation of the last position in the input is used to generate the next token, as is in standard decoding.

\begin{algorithm}[ht]
  \caption{Fast Speculative Prefill}
  \label{alg:speculative_prefill}
  \begin{algorithmic}
    \STATE {\bfseries Input:} Query chunks $\mathbf{X}_{1 \dots K}$, Backbone $M_\theta$, Thinking Block $T_{\phi}$, Compression $C_{\phi}$, Chunk Size $c$
    \STATE $t = 0$
    \STATE $kv = init\_cache()$
    \STATE $\mathbf{S}_{i} = \hat{0}\in\mathbb{R}^{c\times d} \quad i\in[0,\dots,K-1]$ \COMMENT{Speculate thoughts}
    \STATE $S = concat([\mathbf{S}_1, \dots, \mathbf{S}_k])$
    \STATE $X = concat([\mathbf{X}_1, \dots, \mathbf{X}_k])$
    \STATE $finished=false$
    \WHILE{$not\ \ finished$}
    \STATE $\tilde{X} = S + X$
    \STATE $H = M_\theta(\tilde{X}, past\_kv=kv)$
    \STATE $H^{out} = extract(H, L^{out})$
    \STATE $\mathbf{H}_i^{out} = chunk(H^{out}, c) \quad i \in [t, \dots, K]$
    \STATE $\mathbf{S}_{i+1} = \mathcal{C}_{\phi_2}(T_{\phi_1}(\mathbf{H}_i^{out}))$
    \STATE $t = find\_first\_true(\mathbf{S}_i \neq \hat{0})$
    \IF{$t$ is $None$}
    \STATE $finished=true$
    \STATE $kv.cache(H, max\_index=None)$
    \ELSE
    \STATE $kv.cache(H, max\_index=t)$
    \STATE $S = concat([\mathbf{S}_{t}, \hat{0}, \dots, \hat{0}])$
    \STATE $X = concat([\mathbf{X}_{t}, \mathbf{X}_{t+1}, \dots, \mathbf{X}_k])$
    \ENDIF
    \ENDWHILE
    \STATE return $kv, \mathbf{S}_{-1}$
  \end{algorithmic}
\end{algorithm}

\subsection{Constructing Chunk-Level Supervision}\label{app:data-gen}

As mentioned in the main text, Section~\ref{sec-data-gen}, chunk level supervision for the Thinking Block is obtained by mapping reasoning steps to specific positions in the input query. The mapping is realized by placing special indicator tokens in the query text.

Once the query is tokenized, the special tokens allow us to construct a matching reasoning array sharing the same length.
At every position in the query containing an indicator token, we insert the matching reasoning sequence to the same position in the reasoning array. In the vast majority of cases, the desired supervision is obtained by shifting the reasoning array one position to the left, then removing all entries containing a special token in the query sequence, from the query tokens and the reasoning array.

A small number of cases contain multiple consecutive indicator tokens that are placed in succession. 
In these cases, we assign the matching steps in order, starting from the first token position \emph{preceding} the first indicator. In case the reasoning steps \emph{"overflow"} the token sequence length, these steps are concatenated together with a special separator token.

\subsection{State Tracking Tasks}\label{app-state-track}

We provide the definitions for the state tracking tasks used for evaluation in Section~\ref{res-state}, \textit{"Parity"} and \textit{"Variable Assignment"} (abbreviated to \textit{"Vars"}).

\textbf{Parity:} The parity task acts as a natural language version of the standard parity prediction problem over binary sequences, which is commonly used when evaluating extrapolation capabilities in various architectures \cite{dehghani2019universaltransformers, anil2022exploringlengthgeneralizationlarge, delétang2023neuralnetworkschomskyhierarchy}.
Over binary sequences, e.g., $s=00101011$, the task is predicting $sum(s)_{\bmod 2}$.
To represent the task in natural language, we transform the binary sequence into a query describing operations applied to a coin by two entities. 
When providing supervision to \tb, the indicator token mentioned in Section~\ref{sec-data-gen} is placed after each operation, paired with the current state.

An \emph{abbreviated} query is provided in Figure~\ref{fig:inference-view} and a full query, equivalent to the binary sequence $s=1011$, is given here, along with indicator tokens and the matching labels:

\begin{quote}
    \textit{"The coin starts at state heads.\texttt{<T>} Alice doesn't flip the coin.\texttt{<T>} Bob flips the coin.\texttt{<T>} Alice flips the coin.\texttt{<T>} \textcolor{blue}{The final state of the coin is heads.}"}

    \text{Reasoning Targets: }\textit{["heads\texttt{<eos>}", "heads\texttt{<eos>}", "tails\texttt{<eos>}", "heads\texttt{<eos>}"]}
\end{quote}

Where \texttt{<T>} denotes the indicator token and blue denotes text that converted into target tokens for predictions. In this formulation, increasing the number of state operations directly translates to longer binary sequences, i.e., more operations applied to the coin.

\textbf{Variable Assignment (Vars):} This task extends the state tracking setting to multiple integer-valued variables whose updates may depend on one another, requiring complex state updates \cite{anil2022exploringlengthgeneralizationlarge}.
Unlike Parity, which tracks a single binary state, Vars requires maintaining the values of multiple variables (e.g., $a, b, c$) and executing arithmetic operations (e.g., $b = b + a$) that reference the current state of other variables.

The task is represented in natural language as a sequence of assignment operations.
To generate supervision for \tb, similar to the Parity task, we insert the indicator token \texttt{<T>} after each operation paired with the value of the updated variable.

An example query, equivalent to the provided raw input, is formatted below with indicator tokens and the corresponding reasoning targets (derived using modular arithmetic consistent with the final values):

\begin{quote}
    \textit{"Track the variables values: a=1; b=2 a=a+b\texttt{<T>} b=b+a\texttt{<T>} b=b+3\texttt{<T>}\textcolor{blue}{Final values: a=3 b=8}"}

    \text{Reasoning Targets: }\textit{["a=3\texttt{<eos>}", "b=5\texttt{<eos>}", "b=8\texttt{<eos>}"]}
\end{quote}

\subsection{General Reasoning Capabilities}\label{app-gen-reas}

\subsubsection{Coconut - Additional Results}\label{app-coco-res}
As mentioned in the main paper, we scale computation with Coconut by increasing the number of latent thinking tokens. This is achieved by increasing the number of latent tokens introduced in each round of the curriculum, out of 3 rounds in total, excluding the initial phase training the CoT checkpoint used for initialization.
As is evident in Table~\ref{tab:coco-perf}
The experiment in the main paper uses $2$ thinking tokens per round, $6$ in total.

\begin{table}[h!]
    \centering
    \caption{Coconut \cite{hao2024traininglargelanguagemodels} performance and speedup as computation increases, by increasing the number of latent steps.}
    \label{tab:coco-perf}
    \begin{tabular}{crr} 
        \toprule
        Num. Latents & Acc & Speedup \\
        \midrule
        6  & 32.65  & 3.14$\times$ \\
        9  & 33.48  & 2.53$\times$ \\
        12 & 31.29  & 2.08$\times$ \\
        15 & 32.35  & 1.77$\times$ \\
        21 & 32.65  & 1.38$\times$ \\
        \bottomrule
    \end{tabular}
\end{table}

\subsubsection{Data Construction for GSM}\label{app-gsm-data}

As mentioned in the introduction, for the GSM task, we rely on a strong teacher model (Gemini-2.5 Flash \cite{comanici2025gemini}) to transform existing parsed CoT data into the mapping used to supervise the thinking head.
We achieve this by prompting the teacher model with a prompt similar to the one presented in Figure \ref{fig-gsm-data-gen}, removing two of the in context examples for readability.
Using the structured output, we are able to extract the transformed query and matching thinking steps.
To validate the process, we only include samples whose queries and thinking steps provide an exact string match to the source query and thinking steps, once the indicator tokens \texttt{<THINK>} are removed, discarding any samples that result in a mismatch.
After validation we are left with $375,101 / 385,620$ samples, or $97.2\%$.

\begin{figure}[h!]
  \begin{center}
    \centerline{\includegraphics[width=\columnwidth]{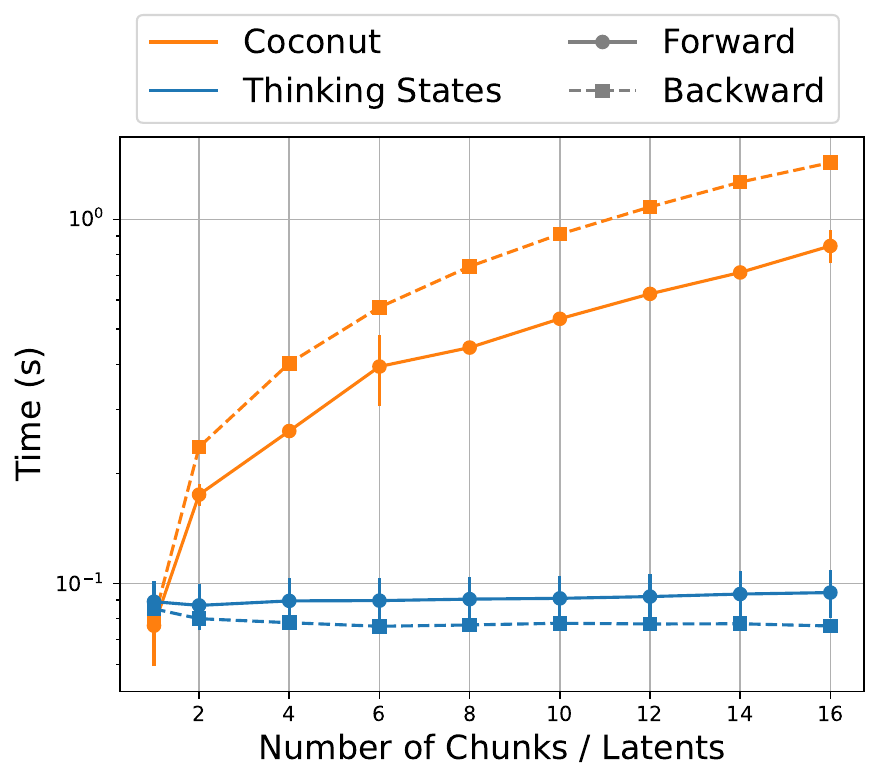}}
    \caption{\textbf{Training Cost Analysis} Comparison of wall-clock time for forward and backward passes on a fixed-length input ($L=128$) as a function of the number of recurrent steps (chunks). Methods relying on BPTT demonstrate linear scaling in time, circumvented by teacher forcing in \arch despite the added dimensionality and computational overhead of \tb.}
    \label{fig:training-compute}
  \end{center}
  \vskip -0.2in
\end{figure}

\subsubsection{Training Compute}\label{app-training-efficiency}

We quantify the computational advantage of training \arch with teacher forcing compared to latent reasoning methods that rely on Backpropagation Through Time (BPTT), such as Coconut. Figure~\ref{fig:training-compute} illustrates the wall-clock time required for a single forward and backward pass on a fixed-length sequence, as a function of the number of recurrent steps (Coconut) or number of chunks\footnote{Here the chunk size accordingly, we note that $k$ chunks imply $k$ recurrent steps at inference time.} (\arch), by controlling the chunk size relative to the sequence length.
We use a batch size of 1 and a sequence length of 128 tokens.

As shown in Figure~\ref{fig:training-compute}, training with BPTT exhibits a linear increase in wall-time for both forward and backward passes as the number of latent steps grows. In contrast, \arch maintains close to constant training time regardless of the number of chunks, as the availability of ground-truth states allows for fully parallelized computation.
The small variation in wall-time for \arch is explained by the additional dimensionality of decoding reasoning \emph{sequence} from fixed-size token chunks.
Yet, as visible in the figure, these variations are small compared to the increase in cost due to BPTT.

\input{data_con_prompt}

%% file: data_con_prompt.tex
\begin{figure*}
\centering

\begin{tcolorbox}[
    colback=gray!5!white,    
    colframe=gray!75!black,  
    title=\textbf{Prompt for GSM Data Generation}, 
    fonttitle=\bfseries,
    enhanced,                
    boxrule=0.8pt,
    width=\linewidth
]
\small 
\ttfamily 

\linespread{0.6}\selectfont   

You are an expert in computational linguistics. Your task is to augment a given query with thinking markers (\texttt{<THINK>}) based on a provided reasoning trace. This involves identifying the precise locations in the text where specific calculations can be performed.

\vspace{0.5em}
\textbf{Task Instructions:}

\begin{enumerate}[leftmargin=*, label=\arabic*., nosep, topsep=0ex, partopsep=0pt]
    \item \textbf{Analyze Inputs:} You will receive two inputs:
    \begin{itemize}[leftmargin=*]
        \item A \texttt{Query}: A question or word problem.
        \item A \texttt{Reasoning Trace}: An ordered list of strings, where each string is a distinct mathematical calculation required to solve the query.
    \end{itemize}

    \item \textbf{Produce Two Outputs:}
    \begin{itemize}[leftmargin=*]
        \item \texttt{query}: This is the original \texttt{Query} text, but with the special token \texttt{<THINK>} inserted at specific locations.
        \item \texttt{thinking}: This is the ordered list of calculations.
    \end{itemize}

    \item \textbf{Rules for \texttt{<THINK>} Placement:}
    \begin{itemize}[leftmargin=*]
        \item The number of \texttt{<THINK>} tokens inserted into the \texttt{query} must be exactly equal to the number of calculations in the \texttt{Reasoning Trace}.
        \item The order of the \texttt{<THINK>} tokens in the \texttt{query} must correspond one-to-one with the order of the calculations in the \texttt{Reasoning Trace}.
        \item For each calculation, you must insert its corresponding \texttt{<THINK>} token at the \textbf{earliest possible location} in the query. This location is defined as the point immediately after the word or phrase that provides the final piece of information needed to perform that specific calculation.
    \end{itemize}

    \item \textbf{Rule for the \texttt{thinking} List:}
    \begin{itemize}[leftmargin=*]
        \item The \texttt{thinking} output list is simply a direct copy of the ordered list of calculations provided in the \texttt{Input Reasoning Trace}.
    \end{itemize}

    \item \textbf{Output Format:}
    \begin{itemize}[leftmargin=*]
        \item The output should be provided as a python dictionary with two keys: \texttt{query} and \texttt{thinking}.
        \item The values should be wrapped in <<>> to support easy parsing.
        \item Example:\\
        \{\\
        "query": "<<Your modified query here>>",\\
        "thinking": <<['<<calculation1>>', '<<calculation2>>', ...]>>\\
        \}
    \end{itemize}
\end{enumerate}

\noindent\rule{\linewidth}{0.4pt} 

\textbf{\small Examples}

\textbf{Example 1:}
\begin{itemize}[leftmargin=*]
    \item \textbf{Input Query:} 'Hannah has three dogs. The first dog eats 1.5 cups of dog food a day. The second dog eats twice as much while the third dog eats 2.5 cups more than the second dog. How many cups of dog food should Hannah prepare in a day for her three dogs?'
    \item \textbf{Input Reasoning Trace:} \texttt{['<<1.5*2=3>>', '<<3+2.5=5.5>>', '<<1.5+3+5.5=10>>']}
\end{itemize}
\textbf{Required Output:}\\
\{\\
"query": <<'Hannah has three dogs. The first dog eats 1.5 cups of dog food a day. The second dog eats twice as much\texttt{<THINK>} while the third dog eats 2.5 cups more than the second dog.\texttt{<THINK>} How many cups of dog food should Hannah prepare in a day for her three dogs?\texttt{<THINK>} '>>,\\
"thinking": <<['<<1.5*2=3>>', '<<3+2.5=5.5>>', '<<1.5+3+5.5=10>>']>>\\
\}

\noindent\rule{\linewidth}{0.4pt}

\textbf{\small Task}

Now, perform this transformation for the following input and return your output as a python dictionary.

\textbf{Query:}\\
\texttt{<<QUERY>>}

\textbf{Reasoning Trace:}\\
\texttt{<<REASONING\_TRACE>>}

\textbf{Produce your output as a python dictionary in the specified format below:}\\
\{\\
"query": <<ADD YOUR OUTPUT HERE>>,\\
"thinking": <<ADD YOUR OUTPUT HERE>>\\
\}

\end{tcolorbox}
\caption{The abbreviated version of few-shot prompt used to generate the alignment data. The prompt instructs the model to act as a computational linguist and inject \texttt{<THINK>} tokens into the query corresponding to specific reasoning steps. The full prompt used is identical except for containing 2 additional in-context examples, similar to the example provided.}
\label{fig-gsm-data-gen}
\end{figure*}
